\documentclass[a4paper,fleqn]{cas-sc}

\usepackage{amsmath,amsfonts}
\usepackage{algorithmic}
\usepackage{array}
\usepackage[caption=false,font=normalsize,labelfont=sf,textfont=sf]{subfig}
\usepackage{textcomp}
\usepackage{stfloats}
\usepackage{url}
\usepackage{verbatim}
\usepackage{graphicx}
\usepackage{lipsum}
\def\BibTeX{{\rm B\kern-.05em{\sc i\kern-.025em b}\kern-.08em
    T\kern-.1667em\lower.7ex\hbox{E}\kern-.125emX}}
\usepackage{balance}

\usepackage{multirow}
\usepackage{arydshln}
\usepackage{colortbl}

\DeclareMathOperator*{\argmax}{arg\,max}

\usepackage{color}

\usepackage[authoryear,longnamesfirst]{natbib}
\usepackage{arydshln}
\def\tsc#1{\csdef{#1}{\textsc{\lowercase{#1}}\xspace}}
\tsc{WGM}
\tsc{QE}
\tsc{EP}
\tsc{PMS}
\tsc{BEC}
\tsc{DE}

\usepackage{ulem}
\usepackage{subcaption}

\begin{document}
\let\WriteBookmarks\relax
\def\floatpagepagefraction{1}
\def\textpagefraction{.001}

\shorttitle{SocialPET for few-shot stance detection in social media}

\shortauthors{P Jamadi Khiabani et~al.}

\title [mode = title]{SocialPET: Socially Informed Pattern Exploiting Training for Few-Shot Stance Detection in Social Media}                      



%
\author[1]{Parisa Jamadi Khiabani}[orcid=0000-0002-0433-5196]

\cormark[1]


\ead{p.jamadikhiabani@qmul.ac.uk}


\credit{Conceptualization, Methodology, Software, Data Curation, Formal Analysis, Writing - Original Draft}

\affiliation[1]{organization={Queen Mary University of London},
    addressline={Mile End Road}, 
    city={London},
    postcode={E1 4NS}, 
    country={United Kingdom}}

\author[1]{Arkaitz Zubiaga}[orcid=0000-0003-4583-3623]


\credit{Conceptualization, Supervision, Writing - Review \& Editing}

%
%

\cortext[cor1]{Corresponding author}



\begin{abstract}
 Stance detection, as the task of determining the viewpoint of a social media post towards a target as `favor' or `against', has been understudied in the challenging yet realistic scenario where there is limited labeled data for a certain target. Our work advances research in few-shot stance detection by introducing SocialPET, a socially informed approach to leveraging language models for the task. Our proposed approach builds on the Pattern Exploiting Training (PET) technique, which addresses classification tasks as cloze questions through the use of language models. To enhance the approach with social awareness, we exploit the social network structure surrounding social media posts. We prove the effectiveness of SocialPET on two stance datasets, Multi-target and P-Stance, outperforming competitive stance detection models as well as the base model, PET, where the labeled instances for the target under study is as few as 100. When we delve into the results, we observe that SocialPET is comparatively strong in identifying instances of the `against' class, where baseline models underperform.\footnote{Our implementation will be publicly available at \url{https://github.com/ParisaJamadi/socialpet}.}
\end{abstract}




\maketitle

\section{Introduction}

Social media platforms offer a goldmine to collect data for analyzing opinions and attitudes expressed by large numbers of users \citep{alturayeif2023enhancing}, which because of the large volume requires development of automated tools to support stance detection from textual content \citep{aldayel2021stance,kuccuk2020stance}. Stance detection is the process of automatically determining a user's viewpoint or position as favor or against regarding a particular subject of interest, often known as the target \citep{alturayeif2023systematic,khiabani2023few}.
In particular, there is a notable interest within the Natural Language Processing (NLP) community for examining the identification of attitudes expressed towards political figures on Twitter \citep{mohammad2016semeval,sobhani2017dataset}.

Much of the previous research in stance detection has generally assumed that there is sufficient training data to develop a model that determines the stance towards a particular target. In a realistic scenario, however, one may have access to limited training data when new targets emerge for which sufficient data could not be labeled. This is also the case where one has a limited budget and cannot afford continually labeling big amounts of data for new targets. This motivates our research looking at the development of a few-shot stance detection methodology that can achieve competitive results when training samples are limited \citep{khiabani2023few,jiang2022few}.

In building a few-shot stance detection method, we propose Socially Informed PET (SocialPET). Our method builds on the well-known Pattern Exploiting Training (PET) \citep{schick2020exploiting}, a technique that uses cloze questions to perform few-shot text classification through the use of patterns that are fed to a language model. Our proposed SocialPET encapsulates community information derived from social media into the process the process of defining the pattern. While PET has proven effective for few-shot text classification, the patterns it uses as cloze questions are not natively designed to exploit the social nature of the stance detection. With SocialPET we propose an enhanced alternative to PET which leverages the structure of the social networks of users to further refine the pattern with the user's community information; our proposed enhancement builds on the assumption of the theory of homophily \citep{mcpherson2001birds}, whereby like-minded users are more likely to be connected to each other in social media. To the best of our knowledge, our proposed SocialPET is in turn the first to ever propose to incorporate insights from the social network structure into the process of defining a socially informed pattern with PET.
 
Through systematic experimentation encompassing two different dataset named P-stance and Multi-target, six target pairs within a few-shot in-target stance detection scenario, our findings robustly demonstrate the efficacy of SocialPET. In comparison, it consistently surpasses the performance of three state-of-the-art in-target stance detection models and a leading pre-trained language model, RoBERTa. Interestingly, SocialPET overcomes limitations of baseline models which are particularly prominent in the `Against', where our socially informed model provides substantial benefits.

\textbf{Contributions.} Our work makes the following novel contributions:
\begin{itemize} 
 \item we propose a novel few-shot stance detection method named SocialPET that infuses socially informed knowledge into the pattern generation process of Pattern Exploiting Training (PET).

 \item by introducing SocialPET, we propose an innovative way of tackling the understudied few-shot stance detection task through a socially informed approach, but also we are the first to propose an extension of the PET model that incorporates a social apapter into it.

 \item we perform experiments on six different targets from two popular stance detection datasets, namely P-Stance and Multi-target, to enable more generalizable analysis.

 \item we delve into the analysis of results, not only looking at overall performance, but also delving into the specific cases for further understanding of the challenges of the few-shot stance detection task and how a socially informed approach can support it.
\end{itemize}

\textbf{Paper structure.} The remainder of the paper is structured as follows. In Section \ref{sec:related-work}, we delve into related work, exploring the challenges associated with stance detection and the utilization of multimodal embeddings for stance detection. \ref{sec:Methodology} introduces our proposed method, SocialPET, while Section \ref{sec:Experiments} outlines the experiment settings. The results of our experiments are detailed in Section \ref{sec:Results}, with a more in-depth discussion provided in Section \ref{sec:discussion}. Finally, we draw conclusions in Section \ref{sec:conclusion} to complete the paper.

\section{Related Work}
\label{sec:related-work}

In this section, we discuss related literature in three key areas of research: we start with stance detection, followed by few- and zero-shot approaches to stance detection, concluding with the incorporation of social features in stance detection.

\subsection{Stance detection}
\label{ssec:stance-classification}

Stance detection research is of widespread interest, with multiple applications to support other tasks such as such as opinion mining \citep{sun2017review}, sentiment analysis \citep{medhat2014sentiment}, determining rumor veracity \citep{kochkina2023evaluating}, and detecting fake news \citep{shu2017fake}. Traditionally, stance detection research has predominantly focused on user-generated content found across various online platforms, including blogs, website comment sections, applications, and social media posts \citep{conforti2020stander}. Notably, Twitter has been a widely utilized data source, primarily due to the ease of accessing its API \citep{zubiaga2016stance,mohammad2016semeval} until the most recent changes in the service. 

Stance detection methodologies vary based on the nature of the text and the specific relationship being described. In instances such as Twitter, the focus is often on identifying the author's stance (for/against/neutral, or often only for/against) towards a proposition or target \citep{mohammad2016semeval}. A shared task organized at SemEval 2016 \citep{mohammad2016semeval} gained substantial popularity as it provided a dataset of tweets categorized by stance, and has become a widely used benchmark dataset for stance detection originates from Twitter \citep{jiang2022few}; it has however become less popular recently due to the aging of the dataset which has led to tweets being unavailable for rehydration \citep{zubiaga2018longitudinal}, reason why we cannot use it in our research. In the first part of the dataset, reflecting a supervised setting, the tweets express opinions on five specific topics: ``Atheism," ``Climate Change," ``Feminist Movement," ``Hillary Clinton," and ``Legalization of Abortion." The second part of the dataset, designed for weakly supervised settings, consists of tweets centered around a single topic.

Recent methods for stance detection employ a variety of linguistic features. These include elements such as word/character n-grams, dependency parse trees, and lexicons \citep{sun2018stance,sridhar2015joint,hasan2013stance}. Furthermore, there is a line of research proposing end-to-end neural network approaches that independently learn topics and opinions, incorporating them through mechanisms like memory networks \citep{mohtarami2018automatic}, bidirectional conditional LSTM \citep{augenstein2016stance}, or neural attention \citep{du2017stance}. Additionally, some neural network approaches leverage lexical features in their models \citep{riedel2017simple,hanselowski2018retrospective}. \cite{saenz2021interpreting} introduced a novel approach utilizing a BERT-based classification model. They complemented this model with an attention-based mechanism aimed at identifying key words that play a crucial role in stance classification. On a similar note, \cite{jiang2022few} proposed an innovative BERT model that integrates knowledge enhancement. This involves incorporating triples from knowledge graphs directly into the text, providing domain-specific knowledge for more informed stance classification.
\cite{li2023new} introduced a novel task called Target-Stance Extraction (TSE) to address stance detection in scenarios where the target is not known in advance, as is often the case in social media texts. They proposed a two-stage framework that first identifies the relevant target in the text and then detects the stance towards the predicted target. \cite{alturayeif2024correction} introduced two Multi-Task Learning (MTL) models, Parallel MTL and Sequential MTL, incorporating sentiment analysis and sarcasm detection tasks to improve stance detection in social media contexts. It explores various task weighting techniques and provides empirical evidence of their effectiveness in MTL models. Through evaluations on benchmark datasets, the proposed Sequential MTL model with hierarchical weighting achieves state-of-the-art results, highlighting the potential of MTL in enhancing stance detection.

Another line of research in stance detection has focused on making models persistence over time, as indeed it has been demonstrated that a model's performance deteriorates when the test data is more distant in time, due to changes in the data as well as in society leading to changes in public opinion \citep{alkhalifa2022capturing}. \cite{alkhalifa2021opinions} looked into increases the temporal persistence of stance detection models by mitigating performance drop with the use of a temporally adaptive stance classifier.

Recent research in natural language processing is increasingly tending towards using pre-trained and large language models \citep{zubiaga2024natural}, and so is the case in stance detection \citep{li2024mitigating}. We follow a similar line of research to the recent trends of leveraging large language models, however in our case we focus on the specific case of few-shot learning, where we are the first to leverage social network structure to improve the potential of language models. We discuss research on few- and zero-shot stance detection next.

\subsection{Few- and zero-shot stance detection}
\label{ Few- and zero-shot}

Zero-shot and few-shot stance detection involve determining the sentiment or position expressed in text concerning a specific target, when there are minimal or no available training resources for that particular target \citep{wen2023zero}. As one of the widely used baseline methods, Xu and colleagues \citep{xu2018cross} introduced CrossNet, an extension of the BiCond model \citep{augenstein2016stance} that incorporated an Aspect Attention Layer. This addition facilitated the identification of domain-specific aspects for cross-target stance inference by employing self-attention to highlight the essential components of a stance-bearing sentence. The architecture of their model comprises four primary layers: embedding layer, context encoding layer, aspect attention layer, and prediction layer. Results indicated that their model surpassed the performance of the BiCond model.

He et al. \citep{he2022infusing} followed a different approach by integrating Wikipedia knowledge to enrich target representations (BERTweet†). This approach signifies a departure from conventional methodologies, showcasing the continual evolution and diversification of techniques in the realm of stance detection research. In another line of work by \citep{wen2023zero}, the researchers employed a conditional generation framework to establish the associations between the semantics of input, target, and label texts in zero-shots and few-shot settings. However, their utilization of the VAST dataset, featuring 5,630 distinct targets, exposes a limitation in the adaptability of their method to other datasets characterized by constrained training resources for diverse targets. Furthermore, the model's lack of applicability to alternative domains, such as social media, stems from its exclusive emphasis on a news-related debate corpus.

In the realm of leveraging large language models for downstream tasks, a notable method in the literature is PET. Developed by Schick et al., PET involves the formulation of pairs comprising cloze question patterns and associated verbalizers. The objective is to capitalize on the knowledge embedded in pretrained language models. Through fine-tuning models for each pattern-verbalizer pair, the authors generate extensive annotated datasets. PET forms the base of our proposed approach SocialPET, which enhances the former by incorporating insights from the social network for stance detection.

JointCL, another approach for zero-shot stance detection, employs stance contrastive learning to generalize stance features for unseen targets using contextual information. It also utilizes a target-aware prototypical graph contrastive learning strategy, using prototypes to extend graph information to unseen targets and model relationships between known and unseen targets. This versatile approach accommodates diverse scenarios for effective stance inference \citep{liang2022jointcl}. Our work incorporates JointCL as one of the baseline methods we compare against.

Despite the dearth of research in few-shot stance detection, it is recently attracting an increasing interest in the community, including approaches that leverage conversational threads for stance detection \citep{li2023improved} and research in other languages such as Chinese \citep{zhao2023c}. These lines of research are however beyond the scope of our work which is focused on the English language and the stance detection task is restricted to a social media post in isolation with its metadata and social networking context, but leaving the need for collecting conversational threads aside.

While existing approaches for stance detection are confined to analyzing the textual content of posts, our research advances this domain by introducing a novel model, SocialPET. Unlike previous methods, SocialPET not only considers text but also incorporates network features into the refinement of the patterns used for pattern exploiting training. We experiment with SocialPET, and compare it with a range of baseline methods including CrossNet, RoBERTa, PET, and JointCL.

\subsection{Social features for stance detection}
\label{ssec:Social featurs }

In recent years, there has been a growing interest in graph analysis, driven by the prevalence of networks in real-world scenarios. Zubiaga et al., focused on classifying social media users based on their stance toward ongoing independence movements in conflicted territories. They adopt a methodology leveraging users' self-reported locations to create expansive datasets for Catalonia, the Basque Country, and Scotland. The analysis revealed the significant impact of homophily, indicating that users predominantly connect with those who share the same national identity. This approach sheds light on the role of social media in reflecting and influencing perspectives within regions experiencing conflicting national identities \citep{zubiaga2019political}. In another words they tried to explore stance classification of users towards the independence movement in their territory rather than stance classification on a text-based basis.  The classifier takes as input a set of users specifically from a given territory \citep{zubiaga2019political}. While in our proposed method, we perfom text-based stance detection using social media features. Ferreyraa et al, took  homophily characteristics of different communities into account for community-detection to automatic generation of Access-Control Lists (ACLs) (friend lists) \citep{ferreyra2022community}. Graphs, also known as networks, serve as representations of information across diverse domains. They find application in fields such as social sciences \citep{goyal2018graph}. While the combination of textual embeddings and graph embeddings has been explored in prior research, it has received limited attention within the domain of stance detection. For instance, in \citep{ostendorff2019enriching}, researchers introduced an approach where they enhanced a BERT transformer by integrating knowledge graph embeddings derived from Wikidata. Their experiments, conducted on a book classification task, demonstrated the superior performance of their method compared to baseline approaches relying solely on text. 

To the best of our knowledge, there has been no prior investigation in few-shot stance detection that leverages social network structure, not least in combination with textual features. Introducing SocialPET in this paper, we outline the first methodology that effectively incorporates it in few-shot stance detection.

\section{Datasets and Extended Data Collection}
\label{Dataset}

We next introduce the datasets we use for our research. For our experiments, we use and extend two existing stance detection datasets, both of which focus on the political domain including politicians as the targets. As the original datasets only provide the textual content of the tweets, we further enrich them for our purposes by collecting the social network structure of the users involved.

\subsection{Process of collection of social network information}

Where the datasets originally only provide tweet texts, we use their associated tweet IDs and user IDs to extend the datasets by collecting user networks (followers and friends) and liked tweets (tweets they liked or favorited from other users):

\begin{itemize}
  \item \textbf{Retrieval of Followers:} Given a user ID as input, we retrieve the complete list of users who follow that user.

  \item \textbf{Retrieval of Friends:} Given a user ID as input, we retrieve the complete list of friends of that user, i.e. all the other users that are followed by the user.

  \item \textbf{Retrieval of Likes:} We retrieve the entire list of tweets liked by each user, given the user ID of the user in question. As we are interested in user networks, rather than tweet networks, we then obtain the user IDs associated with the liked tweets. As a result, we get a list of user IDs liked by a user (more specifically, the list of user IDs whose tweet(s) has liked the user, which we use as a proxy to determine that the user likes the other user).
\end{itemize}

\subsection{Datasets}

Datasets for our study were selected based on their popularity and also their recency, as access to a large proportion of tweets that have not been deleted would be crucial for our purposes of collecting additional network information. P-Stance and Multi-target are both widely-used datasets that satisfy these criteria.

\paragraph{\textbf{P-Stance dataset \citep{li2021p}.}}
 The P-Stance dataset includes tweets annotated for stance around three political figures as the targets: ``Donald Trump," ``Joe Biden," and ``Bernie Sanders". We enrich this dataset by collecting social network information following the procedure above.
 As expected, a small proportion of tweets from the original dataset were no longer available at the time of our collection, due to reasons such as deletion of tweets or deactivation of associated user accounts. Hence, our collection of social network structure led to a slightly reduced collection of data, with a total of 4,212 tweets and the associated social network for the relevant users. 
 
 The label distribution of the resulting dataset is shown in Table \ref{tab:pstance-stats}. While the number of tweets across targets is relatively similar, we observe significant differences in the distributions across labels, where Donald Trump has the highest ratio of against tweets and Bernie Sanders has the highest ratio of favor tweets.

\begin{table}[htb]
 \normalsize
 \begin{center}
  \caption{Statistics of the resulting P-stance dataset.}
  \label{tab:pstance-stats}
  \begin{tabular}{|p{2.5cm}||c|c||c|}
   \hline
   \textbf{Target}  & \textbf{Favor} & \textbf{Against}        & \textbf{Total}  \\
   \hline
   Donald Trump &  519 & 947      & 1466   \\
   \hline
   Joe Biden & 702 & 716   & 1418   \\
   \hline
   Bernie Sanders  & 776 & 553 & 1329 \\
   \hline
  \end{tabular}
 \end{center}
\end{table}

\paragraph{\textbf{Multi-target dataset \citep{sobhani2017dataset}.}}
 The Multi-target dataset comprises 4,455 tweets that have been manually annotated for their stance towards a target. Each of the tweets in the dataset contains more than one target, with an individual annotation for each of the targets. Given that, in our case, we are only looking at a single target at a time, we duplicate the tweets with each of the targets, e.g. if a tweet has label1 for target1 and label2 for target2, we create two instances with the same text, \{tweet, label1, target1\} and \{tweet, label2, target2\}.
 
 Originally, this dataset comprised tweets centered around four distinct political figures: ``Donald Trump," ``Bernie Sanders," ``Hillary Clinton," and ``Ted Cruz". We chose to exclude the ``Ted Cruz" target from our experiments due to its relatively smaller sample size, making it unsuitable for the few-shot stance detection task, especially in scenarios requiring higher shot counts like 300 and 400. Additionally, since our focus is on 2-class stance detection, we also omitted the samples labeled as ``None" stance, for consistency with the P-Stance dataset.

The resulting dataset statistics are shown in the Table \ref{tab:multi-stats}. As with the P-Stance dataset, we observe different in labels distributions across targets, with Hillary Clinton showing the highest imbalances (toward favor and toward against, respectively). In contrast, Bernie Sanders and Donald Trump tweets show a relatively balanced distribution between favor and against tweets.

\begin{table}[htb]
 \normalsize
 \begin{center}
  \caption{The statistics of the resulting Multi-target dataset.}\label{tab:multi-stats}
  \begin{tabular}{|p{2.5cm}||c|c||c|}
   \hline
   \textbf{Target}  & \textbf{Favor} & \textbf{Against}        & \textbf{Total}  \\
   \hline
   Donald Trump &  699 & 503      & 1202   \\
   \hline
   Hillary Clinton & 331 & 872   & 1203   \\
   \hline
   Bernie Sanders  & 387 & 226 & 613 \\
   \hline
  \end{tabular}
 \end{center}
\end{table}


\section{Methodology}
\label{sec:Methodology}

This section outlines the framework of our proposed SocialPET model.
In Section \ref{ssec:formulation}, we start by formulating the stance detection task. Following that, our proposed SocialPET framework is detailed in its implementation (Section \ref{ssec:social-pet}).

\subsection{Problem Formulation}
\label{ssec:formulation}

The stance detection task consists in categorizing each post $p$ in a collection $P$ by their stance towards each of the targets $t_j$ in a collection of targets $T$. Based on the assumption that each post $p_i$ expresses a viewpoint toward a particular target (denoted as $t$), the objective of the stance detection is to classify a post $p_i$ into one of two stances: $S = \{favor, against\}$. The stance detection model is built from labeled training data $P_{train}$, which we then evaluate on a held-out test set $P_{test}$. In the specific case of few-shot stance detection, the stance detection model is built from a $P_{train}$ training set which only contains a small sample of labeled instances.

In our particular case, we set up the few-shot experiments where $P_{train}$ ranges from 100 to 400 samples in steps of 100. We perform experiments across different targets $t_j$, which all together enable us to perform more generalizable experiments beyond just a single target.

\subsection{The Base Method: Pattern Exploiting Training (PET)}
\label{ssec:pet}

The PET method employs a pattern-verbalizer pair (PVP) approach to solve NLP tasks as cloze questions. Each PVP consists of a pattern, which generates cloze-style questions from input sequences, and a verbalizer, which maps labels to vocabulary words. The model then predicts the most likely label for a given input by filling in the cloze question. During training, the model is fine-tuned using small labeled datasets and auxiliary language modeling to prevent catastrophic forgetting. Multiple PVPs are combined using an ensemble approach, and the resulting soft-labeled dataset is used to train a final classifier. This process allows PET to effectively leverage both labeled and unlabeled data to achieve superior performance in NLP tasks, even with limited training examples \citep{schick2020exploiting}. 
For our experiments, we have formulated our customized patterns and verbalizers for our few-shot stance detection task. In our case, each input consists of the text itself (a), and the name of the target being mentioned in the text (b). To denote boundaries between these two elements, we utilize a pair of vertical bars (||). Each example, represented as x = (a, b), comprises a Comment (a) regarding a Target(b). The objective is to detect whether the author of the comment `a' in favor (1) or against (0) the target `b'. Inspired by the settings of the related X-Stance task presented in \cite{schick2020exploiting}, we employ a combination of two patterns, as follows. Additionally, we establish an English verbalizer, vEn, which translates FAVOR to ``Yes" and AGAINST to ``No".
\[
P1(x) = \text{``}a\text{''} || \uline{\hspace{0.8cm}} \text{``}b\text{''}
\]
\[
P2(x) = a || \uline{\hspace{0.8cm}} b
\]

\subsection{Proposed Method: Socially Informed PET (SocialPET)}
\label{ssec:social-pet}

\begin{figure}[htb]
	\centering
		\includegraphics[scale=.45]{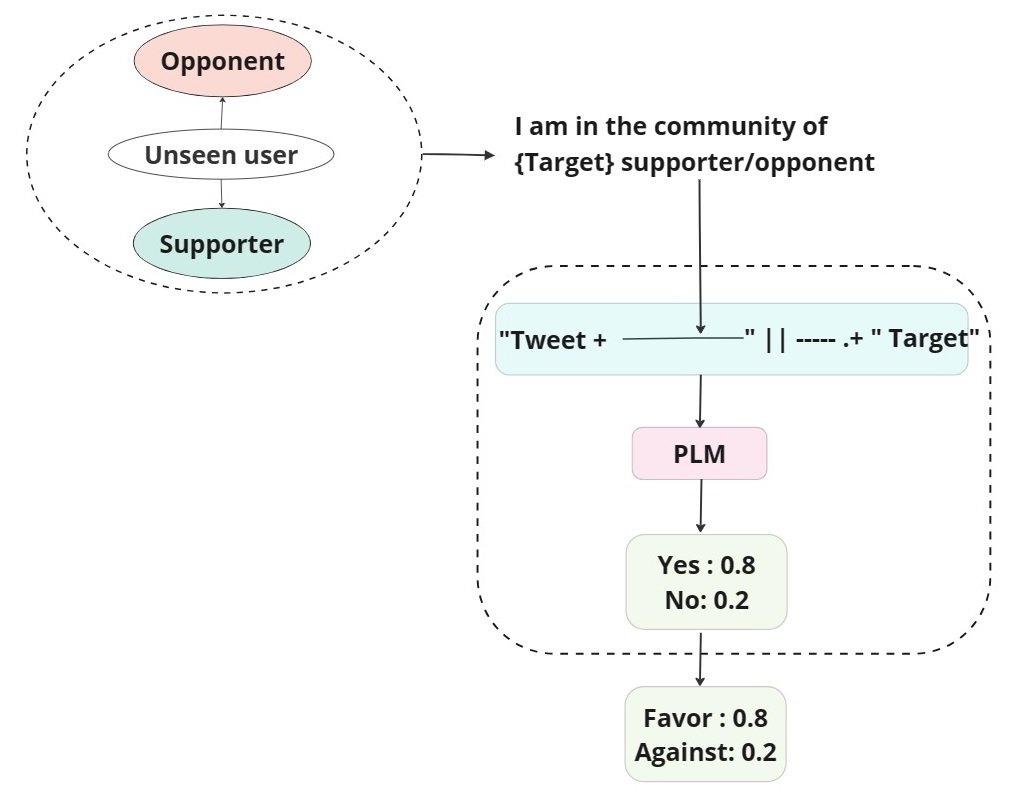}
	\caption{Architecture of the proposed model, SocialPET.}
	\label{fig:architecture}
\end{figure}

Our proposed SocialPET architecture enriches the original capabilities of PET (Pattern Exploiting Training) by incorporating insights derived from the social network into the pattern creation process through a social adapter component that we define (see Figure \ref{fig:architecture}).

The social adapter component enables extending PET's original pattern by adding social awareness to it. More specifically, the social adapter appends an additional sentence at the end of the pattern indicating the community that the user of a specific tweet is predicted to belong to. For example, if a user is predicted to belong to the community of Donald Trump supporters based on their surrounding social network, this information would be appended to the pattern to support the prediction process through SocialPET. Hence, rather than having a pattern that solely relies on the text of a tweet, it incorporates predicted community information based on the social network into the pattern. In order for this approach to be effective, the community prediction strategy needs to carefully defined.

SocialPET's social adapter component follows these criteria to incorporate community information into the pattern:

\begin{itemize}
 \item For tweets in the training data (i.e. those pertaining to the source targets), the community of the user is directly inferred from the label of the tweet in question. For a tweet labeled $L1$, we assume the user belong to community $C_{L1}$, for a tweet labeled $L2$, the user belongs to community $C_{L2}$, etc. For example, if a tweet whose target is Joe Biden is labeled as `support', we consider that this user belongs to the community of Joe Biden supporters.

 \item For tweets in the test data, stance labels are not accessible, and therefore we predict their community following these steps:

 \begin{enumerate}
  \item We use the few shots we incorporate into the training from the destination target to learn about the characteristics of the two associated communities. For the users associated with these few shots, we aggregate all their network features including likes, friends and followers ($N$). As a result, we get aggregated representations of all supporters ($N_S$) and of all opponents ($N_O$) of the destination target, derived from the few shots (i.e. a list of all the users liked, followed and are followed by in the group of supporters and in the group of opponents).

  \item For each tweet in the set, whose label is not known, we aggregate the network information of the associated user ($N_j$), which again includes likes, friends and followers. Once we have this vector with network information, we check which of the two communities linked to the target destination (i.e. supporters $N_S$ or opponents $N_O$) the user in question is closer to. We do this by measuring the overlap between the user and each of the two communities. The one with the highest degree of overlap is the predicted community, i.e.:

  \begin{equation}
    \argmax_{G=\{S, O\}} (N_j \cap N_G) 
  \end{equation}

  where G belongs to a group (supporters or opponents), j is the user being considered.
 \end{enumerate}
\end{itemize}

Once the communities of each of the users $C_j$ in the dataset has been derived through the above two means, we revise the pattern fed into SocialPET to incorporate community-centric social awareness, by appending the phrase \textit{`I am in the community of {$C_j$}'}, as follows:

\[
P1(x) = \text{``}a + \text{I am in the community of } \text{\{$C_j$}\}\text{''} || \uline{\hspace{0.8cm}} \text{``}b\text{''}
\]
\[
P2(x) = a + \text{I am in the community of } \text{\{$C_j$}\}|| \uline{\hspace{0.8cm}} b
\]

Where $C_j$ = {`Bernie Sanders supporter', `Bernie Sanders opponent', `Donald Trump supporter', `Donald Trump opponent', ...}

\section{Experiments}
\label{sec:Experiments}

We next discuss the baseline methods used in our research, lay out the experiment settings and the evaluation metrics we use.

\subsection{Baseline Methods}
\label{Baseline Methods}

To evaluate the effectiveness of our proposed SocialPET methodology, we compare it with a series of strong baselines as well as the widely-used Transformer model RoBERTa.

\begin{itemize}
 \item \textbf{CrossNet \citep{xu2018cross}:} This model is an enhanced version of BiCond method (neural network-based method) powered by a self-attention layer to capture important words
in the input text.
 \item \textbf{RoBERTa \citep{liu2019roberta}:} This method adapts a pre-trained BERT model to handle stance detection tasks. It does this by structuring the input data in a specific format, where the target of interest is placed at the beginning of the input followed by the context, separated by special tokens. This approach allows the model to effectively process and understand both the target and its surrounding context.
 \item \textbf{PET \citep{schick2020exploiting}:} This approach for few-shot text classification and natural language inference leverages pretrained language models (PLMs) like BERT by reformulating input examples into cloze-style phrases. PET consists of three steps: finetuning separate PLMs on small training sets using patterns, annotating unlabeled data with soft labels using the ensemble of PLMs, and training a standard classifier on the soft-labeled dataset. 
 \item \textbf{JointCl \citep{liang2022jointcl}:} This framework leverages joint contrastive learning. It combines stance contrastive learning and target-aware prototypical graph contrastive learning to generalize stance features to unseen targets.
\end{itemize}

\subsection{Experiment Settings}
\label{Experiment Settings}

We perform experiments across various n-shot settings, where $n = \{100, 200, 300, 400\}$. In each of these settings, the training data includes all but the destination target of the dataset in question, in addition to $n$ samples incoroporated into the training from the destination target. This means that for each of the datasets with three targets ($t_1$, $t_2$, $t_3$), we run three different experiments where each of the targets $t_i$ is considered the destination target. For each of the experiments, we perform five runs with five different random seeds:  24, 524, 1024, 1524, and 2024. We then average the performance results across these five runs as the final score that we report.

\paragraph{\textbf{Model Hyperparameters.}}
For our experiments across both datasets, we use RoBERTa. Our hyperparameter choices are influenced by insights from PET research and practical considerations. This includes setting a learning rate of $1e^{-5}$, a batch size of 16, and a maximum sequence length of 256. Furthermore, we implemented the weighted variant of PET with auxiliary language modeling.

\subsection{Evaluation Metrics}

In line with previous research in stance detection \citep{augenstein2016stance, xu2018cross, liang2022jointcl, schick2020exploiting}, we also adopt the macro-averaged F1 score ($MacF_{avg}$) as the main metric to evaluate the performance in our experiments. This enables us to account for the imbalance of labels for some of the targets, enabling a fairer evaluation where we aim to optimize for both labels rather than the majority label only. In our case with binary classification involving the FAVOR and AGAINST classes, the resulting metric is the arithmetic mean of the F1 scores for each class, as follows:

\begin{equation}
 MacF_{avg} = \frac{F1_{favor} + F1_{against}}{2}
\end{equation}

where each of $F1_{favor}$ and $F1_{against}$ is defined as follows:

\begin{equation}
 F1_c = \frac{2 * precision_c * recall_c}{precision_c + recall_c}
\end{equation}

where $c = \{favor, against\}$.

\section{Results}
\label{sec:Results}

We next discuss the results of our experiments on both datasets, P-Stance and Multi-target.

\subsection{SocialPET vs Baselines}

\begin{table*}[htb]
 \normalsize
 \begin{center}
  \caption{Macro-averaged F1 scores for SocialPET vs baseline models for 100-400 shots.}
  \label{tab:socialpet-vs-baselines}
  \begin{tabular}{|l||ccc|ccc||c|}
   \cline{2-8}
   \multicolumn{1}{l|}{Dataset}& \multicolumn{3}{c|}{Multi-target} & \multicolumn{3}{c||}{P-Stance} & Average\\
   \multicolumn{1}{l|}{Target}   & Trump & Sanders   & Clinton   & Trump & Sanders   & Biden &                          \\
   \hline
   \multicolumn{8}{c}{100 shots} \\
   \hline
   \hline
   CrossNet   & 0.57 & 0.37 & 0.53 & 0.5 & 0.41 & 0.63 & 0.5 \\
   RoBERTa  & 0.7 & 0.41 & \textbf{0.54} & 0.5 & 0.54  & \textbf{0.73} & 0.57 \\
   PET   & 0.75 & 0.4 & 0.31 & 0.52 & \textbf{0.57} & 0.67 & 0.54 \\
   JointCl   & 0.75 & 0.38 & 0.53 & 0.41 & 0.3 & 0.43 & 0.47 \\
   \hdashline
   SocialPET     & \textbf{0.76} & \textbf{0.46}  & 0.43 & \textbf{0.75} & 0.54 & 0.71 & \textbf{0.61} \\
   \hline
   \multicolumn{8}{c}{200 shots} \\
   \hline
   CrossNet   & 0.65 & 0.46 & 0.54 & 0.54 & 0.45 & 0.63 & 0.55 \\
   RoBERTa  & 0.75 & 0.46 & 0.58 & 0.65 & \textbf{0.67}  & \textbf{0.76} & 0.65 \\
   PET   & 0.75 & 0.38 & 0.43 & 0.64 & 0.61 & 0.74 & 0.54 \\
   JointCl   & 0.76 & 0.36 & 0.5 & 0.41 & 0.4 & 0.63 & 0.51 \\
   \hdashline
   SocialPET     & \textbf{0.77} & \textbf{0.65}  & \textbf{0.82} & \textbf{0.83} & 0.63 & \textbf{0.76} & \textbf{0.74} \\
   \hline
   \multicolumn{8}{c}{300 shots} \\
   \hline
   CrossNet   & 0.59 & 0.38 & 0.57 & 0.48 & 0.45 & 0.55 & 0.5 \\
   RoBERTa  & 0.75 & 0.55 & 0.63 & 0.68 & 0.68  & \textbf{0.79} & 0.62 \\
   PET   & 0.73 & 0.45 & 0.42 & 0.65 & 0.68 & 0.78 & 0.62 \\
   JointCl   & \textbf{0.76} & 0.54 & 0.44 & 0.42 & 0.59 & 0.69 & 0.57 \\
   \hdashline
   SocialPET     & \textbf{0.76} & \textbf{0.64}  & \textbf{0.7} & \textbf{0.83} & \textbf{0.72} & 0.75 & \textbf{0.73} \\
   \hline
   \multicolumn{8}{c}{400 shots} \\
   \hline
   CrossNet   & 0.69 & 0.35 & 0.57 & 0.58 & 0.53 & 0.65 & 0.56 \\
   RoBERTa  & 0.74 & 0.54 & 0.62 & 0.69 & 0.7  & \textbf{0.8} & 0.68 \\
   PET   & 0.74 & 0.56 & 0.55 & 0.67 & 0.7 & 0.78 & 0.66 \\
   JointCl   & 0.75 & 0.55 & 0.45 & 0.62 & \textbf{0.73} & 0.59 & 0.62 \\
   \hdashline
   SocialPET     & \textbf{0.77} & \textbf{0.58}  & \textbf{0.7} & \textbf{0.84} & \textbf{0.73} & 0.74 & \textbf{0.73} \\
   \hline
  \end{tabular}
 \end{center}
\end{table*}

The results of the comparative analysis between our proposed SocialPET and baseline models across the different n-shot settings are detailed in Table \ref{tab:socialpet-vs-baselines}, for results ranging from 100 to 400 shots.

Results show that, overall, SocialPET is consistently the best when we look at average performances, regardless of the number of shots used. It is also interesting that this overall improvement happens already with as little as 100 shots, even when we compare it with the base model PET. This indicates that the social adapter we add in SocialPET starts to perform well when even a small number of samples is available for making community predictions to generate the socially informed pattern.

Despite the average outperformance of SocialPET with as little as 100 shots, this is not as consistent across the different targets for this small number of samples. We observe that the text-based RoBERTa model shows competitive and at times better performance for two of the targets. However, RoBERTa's performance fades as we increase the number of shots, leading to more consistent improvement of SocialPET over RoBERTa.

Starting from 200 shots, SocialPET quite consistently outperforms other baseline models across targets, not only on average. The exception to this are the Joe Biden target (200-400 shots) and Bernie Sanders (200 shots), where SocialPET underperforms RoBERTa and, in some cases, the base model PET. This may have to do with the social network information of the Biden supporters and opponents which may not be as distinguishable as for the other targets.

When we look at the baseline models designed for stance detection, CrossNet and JointCL, they are both consistently outperformed by our model SocialPET, with the only exception of Hillary Clinton with 100 shots, where the baseline models perform better. On average, however, SocialPET is clearly superior to these two baseline models, with absolute improvements ranging between 10-20\% in most cases.

We next drill down into the performance of SocialPET by looking at performances by class.

\subsection{Analysis of Results by Class}

Having seen the competitive and consistent performance of SocialPET, we next look at how these improvements are distributed across both classes. In the interest of focus, we compare our proposed SocialPET with the base model PET, hence delving into the analysis of the impact of the social adapter on performance. To analyze this, we look at confusion matrices across all datasets and targets in Table \ref{tab:Confusion matrices}. Confusion matrices show correct predictions in the sinister diagonal, and mispredictions in the dexter diagonal. In our analysis, we focus primarily on the dexter diagonals to look at model errors. We highlight in blue the cases where SocialPET makes fewer mistakes, and in red the cases where SocialPET makes more mistakes than PET.

If we look at the ability of SocialPET to improve across the two classes, we observe this particularly prominent in the cases of Donald Trump on P-Stance and Hillary Clinton as targets. In these cases, improvement happens by reducing the number of misclassifications in both directions. Still, the largest improvement in misclassifications is by reducing the number of cases that are actually `Against' and are classified as `Favor'. PET makes large numbers of mistakes in these cases (186 and 149), which SocialPET reduces (61 and 79 respectively).

In three other cases, namely the two Sanders targets and Trump on Multi-target, the improvement reducing misclassifications only happens in one of the two directions. For Donald Trump on Multi-target, SocialPET gets favor-to-against mistakes down from 123 to 114, although the against-to-favor mistakes go up from 60 to 68. These differences are still much more modest than in other cases. For the two Sanders targets, we observe a similar pattern, showing substantial reductions on against-to-favor mistakes (119 to 76 and 74 to 24), but this comes with an increase in favot-to-against mistakes (111 to 206 and 2 to 77). On the positive side, however, SocialPET achieves remarkable improvements in correct Against predictions, going from 217 to 260 and from 1 to 51.

The more negative case, however, as with the overall experiment results presented above, is with the Joe Biden target. In this case, the increase in mistakes happens in both directions against-to-favor (117 to 133) and favor-to-against (92 to 128). Again this indicates that the social network structure in this particular case may not be as informative as with other target, and hence performance of SocialPET is not as good.

\begin{table}[htbp]
\caption{Confusion matrices of SocialPET model vs. PET model for P-stance and Multi-target datasets}
\label{tab:Confusion matrices}
\addtolength{\tabcolsep}{-1pt}
\normalsize
\begin{tabular}{|c|c|c|c|c|c|c|}
\multicolumn{3}{c}{\textbf{P-stance}} & \multicolumn{1}{c}{} & \multicolumn{3}{c}{\textbf{Multi-target}} \\
\cline{1-3}\cline{5-7}
\multicolumn{3}{|c|}{\textbf{Trump}} & \multicolumn{1}{c|}{} & \multicolumn{3}{c|}{\textbf{Trump}} \\
\cline{1-3}\cline{5-7}
\multicolumn{3}{|c|}{\textit{SocialPET}} & \multicolumn{1}{c|}{} & \multicolumn{3}{c|}{\textit{SocialPET}} \\
\cline{1-3}\cline{5-7}
  & Against       & Favor &  &    & Against       & Favor \\
\cline{1-3}\cline{5-7}
Against & 649           & \textbf{\textcolor{blue}{61}}    &  & Against & 229           & \textbf{\textcolor{red}{68}}    \\
Favor   & \textbf{\textcolor{blue}{97}}            & 207   &  & Favor   & \textbf{\textcolor{blue}{114}}            & 373   \\
\cline{1-3}\cline{5-7}
\multicolumn{3}{|c|}{\textit{PET}} & \multicolumn{1}{c|}{} & \multicolumn{3}{c|}{\textit{PET}} \\
\cline{1-3}\cline{5-7}
  & Against       & Favor &  &    & Against       & Favor \\
\cline{1-3}\cline{5-7}
Against & 524            & 186   &  & Against & 237            & 60   \\
Favor   & 116            & 188   &  & Favor   & 123            & 364   \\
\cline{1-3}\cline{5-7}
\multicolumn{7}{c}{} \\
\multicolumn{3}{c}{\textbf{Sanders}} & \multicolumn{1}{c}{} & \multicolumn{3}{c}{\textbf{Sanders}} \\
\cline{1-3}\cline{5-7}
\multicolumn{3}{|c|}{\textit{SocialPET}} & \multicolumn{1}{c|}{} & \multicolumn{3}{c|}{\textit{SocialPET}} \\
\cline{1-3}\cline{5-7}
  & Against       & Favor &  &    & Against       & Favor \\
\cline{1-3}\cline{5-7}
Against & 260           & \textbf{\textcolor{blue}{76}}     &  & Against & 51           & \textbf{\textcolor{blue}{24}}    \\
Favor   & \textbf{\textcolor{red}{206}}            & 346   &  & Favor   & \textbf{\textcolor{red}{77}}            & 157   \\
\cline{1-3}\cline{5-7}
\multicolumn{3}{|c|}{\textit{PET}} & \multicolumn{1}{c|}{} & \multicolumn{3}{c|}{\textit{PET}} \\
\cline{1-3}\cline{5-7}
  & Against       & Favor &  &    & Against       & Favor \\
\cline{1-3}\cline{5-7}
Against & 217           & 119    &  & Against & 1           & 74    \\
Favor   & 111           & 441    &  & Favor   & 2            & 232   \\
\cline{1-3}\cline{5-7}
\multicolumn{7}{c}{} \\
\multicolumn{3}{c}{\textbf{Biden}} & \multicolumn{1}{c}{} & \multicolumn{3}{c}{\textbf{Clinton}} \\
\cline{1-3}\cline{5-7}
\multicolumn{3}{|c|}{\textit{SocialPET}} & \multicolumn{1}{c|}{} & \multicolumn{3}{c|}{\textit{SocialPET}} \\
\cline{1-3}\cline{5-7}
  & Against       & Favor &  &    & Against       & Favor \\
\cline{1-3}\cline{5-7}
Against & 366           & \textbf{\textcolor{red}{133}}    &  & Against & 233           & \textbf{\textcolor{blue}{79}}    \\
Favor   & \textbf{\textcolor{red}{128}}            & 354   &  & Favor   & \textbf{\textcolor{blue}{18}}            & 75   \\
\cline{1-3}\cline{5-7}
\multicolumn{3}{|c|}{\textit{PET}} & \multicolumn{1}{c|}{} & \multicolumn{3}{c|}{\textit{PET}} \\
\cline{1-3}\cline{5-7}
  & Against       & Favor &  &    & Against       & Favor \\
\cline{1-3}\cline{5-7}
Against & 382           & 117    &  & Against & 163            & 149   \\
Favor   & 92           & 390    &  & Favor   & 34            & 59   \\
\cline{1-3}\cline{5-7}
\end{tabular}
\label{tab:confusion-matrices}
\end{table}

We further look at the aggregated confusion matrices for each dataset in Table \ref{tab:Confusion-matrices-aggregated}, where we aggregate correct predictions and mispredictions in each dataset. These results confirm some of the findings from the analysis above. When it comes to correct predictions, improvement comes in the `Against' class, with an increase from 1123 to 1275 in the P-Stance dataset, and an increase from 401 to 513 in the Multi-target dataset. Improvement in the reduction of misclassifications happens in the against-to-favor cases, dropping from 422 to 270 in P-stance and from 283 to 171 in Multi-target.

\begin{table}[htbp]
\caption{Aggregate confusion matrix combining predictions across all three targets for each of the P-stance and Multi-target datasets}
\label{tab:Confusion-matrices-aggregated}
\addtolength{\tabcolsep}{-1pt}
\normalsize
\begin{tabular}{|c|c|c|c|c|c|c|}
\multicolumn{3}{c}{\textbf{P-stance}} & \multicolumn{1}{c}{} & \multicolumn{3}{c}{\textbf{Multi-target}} \\
\cline{1-3}\cline{5-7}
\cline{1-3}\cline{5-7}
\multicolumn{3}{|c|}{\textit{SocialPET}} & \multicolumn{1}{c|}{} & \multicolumn{3}{c|}{\textit{SocialPET}} \\
\cline{1-3}\cline{5-7}
  & Against       & Favor &  &    & Against       & Favor \\
\cline{1-3}\cline{5-7}
Against & 1275           & \textbf{\textcolor{blue}{270}}    &  & Against & 513           & \textbf{\textcolor{blue}{171}}    \\
Favor   & \textbf{\textcolor{red}{431}}            & 907   &  & Favor   & \textbf{\textcolor{red}{209}}            & 605   \\
\cline{1-3}\cline{5-7}
\multicolumn{3}{|c|}{\textit{PET}} & \multicolumn{1}{c|}{} & \multicolumn{3}{c|}{\textit{PET}} \\
\cline{1-3}\cline{5-7}
  & Against       & Favor &  &    & Against       & Favor \\
\cline{1-3}\cline{5-7}
Against & 1123            & 422   &  & Against & 401            & 283   \\
Favor   & 319            & 1019   &  & Favor   & 159            & 655   \\
\cline{1-3}\cline{5-7}
\multicolumn{7}{c}{} \\
\end{tabular}
\end{table}

\section{Discussion}
\label{sec:discussion}

Our experiments testing our proposed SocialPET model across six targets from two different datasets show state-of-the-art performance of our model compared to a range of competitive baselines. Beyond this, a closer look at the results brings useful insights into the challenges of the stance detection task.

All in all, our results show an overall performance improvement of SocialPET over PET, achieving a better balance of predictions across both classes leading to better aggregated performance, with the performance improvement being particularly noticeable in the `Against' class. Our results suggest that, while text-based basline methods such as PET can perform well on the `Favor' class, use of social network structure provides complementary information to support improvement on identifying instances of the `Against' class.

While overall improvements are quite consistent, we do still observe a negative case with Joe Biden as the target, where the model underperforms the text-based base model PET. Indeed, not every case presents an ideal social network structure that supports the stance detection task, as we see with Joe Biden.

\begin{figure}[htb]
	\centering
		\includegraphics[scale=.45]{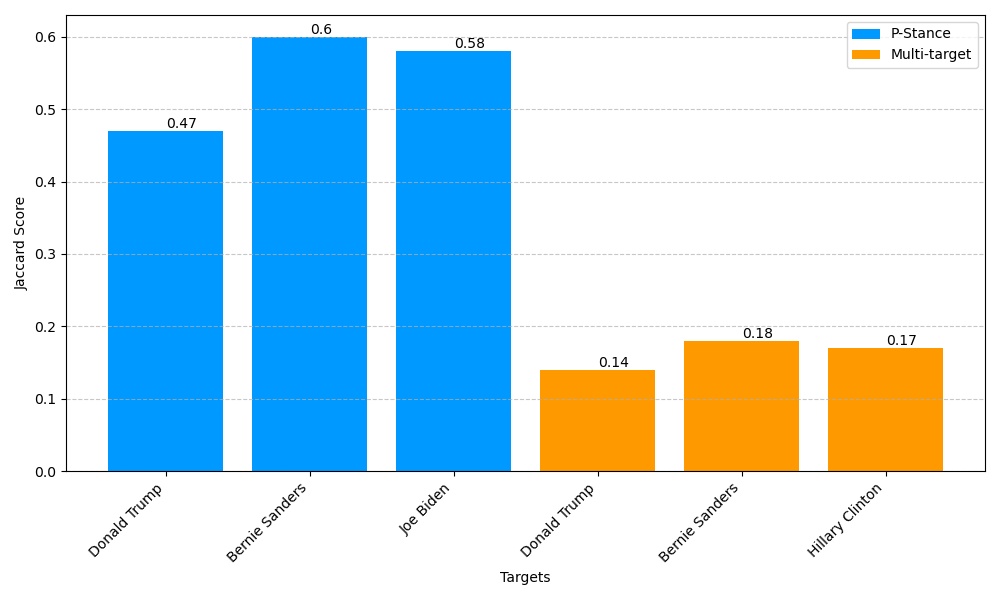}
	\caption{Jaccard Scores of Supporters and Opponents Groups for Each Target for P-Stance and Multi-target Datasets.}
	\label{fig:jaccard_score}
\end{figure}

To better understand the variations in performance across different targets, we look at the extent to which the social networks of supporters and opponents of each target resemble. In Figure \ref{fig:jaccard_score}, we show the Jaccard scores measuring the similarity between the aggregated social network of supporters against the aggregated social network of opponents of each target. Higher Jaccard scores indicate higher overlap between the social networks of supporters and opponents of that target.

We observe big differences between the Jaccard scores for each of the datasets, where the scores for Multi-target range from 0.14-0.18 and the scores for P-Stance range from 0.47-0.6. This is however likely due to how each of the datasets was collected and sampled that ultimately led to more or less overlaps between the networks of both groups. Interestingly, however, we observe that, for both datasets, the target with a smaller Jaccard score is Donald Trump; this indicates that, among the targets under study, Donald Trump is the one whose supporters and opponents differ the most in their social network, in turn indicating that the information we can derive from their networks for SocialPET is more distinctive and informative. Conversely, we observe that both Joe Biden and Bernie Sanders have comparatively higher Jaccard scores with respect to Donald Trump in the P-Stance dataset; these are in fact the two cases where SocialPET does not perform as competitively, as both Joe Biden and Bernie Sanders are the cases where SocialPET does not consistently outperform other baselines. Most importantly, despite the overall competitive performance of SocialPET, we see that one can estimate and predict the potential of SocialPET to succeed by looking at the overlap of social networks of supporters and opponents of a target.

\section{Conclusion}
\label{sec:conclusion}

In this study, we propose a novel model SocialPET, which building on the widely used Pattern Exploiting Training (PET) model enables stance detection from social media posts by injecting socially informed knowledge into the pattern generation process. With experiments on six different targets across two popular stance detection datasets, P-Stance and Multi-target, we show the overall competitive performance of SocialPET over a range of baseline models, including the base model PET. Interestingly, however, this improvement is particularly noticeable in the `Against' class, where the socially informed SocialPET can complement other techniques.

Despite the overall improvements, there are cases where it is more modest or on occasions negative. While SocialPET has proven overall successful, we aim to look at alternative ways of prioritiing textual vs social input into the model depending on the case, for example by analyzing how useful the social network structure can be in a particular caase in advance.

\section*{Acknowledgments}
Parisa Jamadi Khiabani is funded by the Islamic Development Bank (IsDB). We thank the authors of the P-stance and Multi-target datasets for kindly providing us with the tweet IDs which enables us to complement the datasets.

\printcredits

\bibliographystyle{cas-model2-names}

\bibliography{cas-refs}

\begin{thebibliography}{44}
\expandafter\ifx\csname natexlab\endcsname\relax\def\natexlab#1{#1}\fi
\providecommand{\url}[1]{\texttt{#1}}
\providecommand{\href}[2]{#2}
\providecommand{\path}[1]{#1}
\providecommand{\DOIprefix}{doi:}
\providecommand{\ArXivprefix}{arXiv:}
\providecommand{\URLprefix}{URL: }
\providecommand{\Pubmedprefix}{pmid:}
\providecommand{\doi}[1]{\href{http://dx.doi.org/#1}{\path{#1}}}
\providecommand{\Pubmed}[1]{\href{pmid:#1}{\path{#1}}}
\providecommand{\bibinfo}[2]{#2}
\ifx\xfnm\relax \def\xfnm[#1]{\unskip,\space#1}\fi
\bibitem[{AlDayel and Magdy(2021)}]{aldayel2021stance}
\bibinfo{author}{AlDayel, A.}, \bibinfo{author}{Magdy, W.}, \bibinfo{year}{2021}.
\newblock \bibinfo{title}{Stance detection on social media: State of the art and trends}.
\newblock \bibinfo{journal}{Information Processing \& Management} \bibinfo{volume}{58}, \bibinfo{pages}{102597}.
\bibitem[{Alkhalifa et~al.(2021)Alkhalifa, Kochkina and Zubiaga}]{alkhalifa2021opinions}
\bibinfo{author}{Alkhalifa, R.}, \bibinfo{author}{Kochkina, E.}, \bibinfo{author}{Zubiaga, A.}, \bibinfo{year}{2021}.
\newblock \bibinfo{title}{Opinions are made to be changed: Temporally adaptive stance classification}, in: \bibinfo{booktitle}{Proceedings of the 2021 workshop on open challenges in online social networks}, pp. \bibinfo{pages}{27--32}.
\bibitem[{Alkhalifa and Zubiaga(2022)}]{alkhalifa2022capturing}
\bibinfo{author}{Alkhalifa, R.}, \bibinfo{author}{Zubiaga, A.}, \bibinfo{year}{2022}.
\newblock \bibinfo{title}{Capturing stance dynamics in social media: open challenges and research directions}.
\newblock \bibinfo{journal}{International Journal of Digital Humanities} \bibinfo{volume}{3}, \bibinfo{pages}{115--135}.
\bibitem[{Alturayeif et~al.(2023a)Alturayeif, Luqman and Ahmed}]{alturayeif2023enhancing}
\bibinfo{author}{Alturayeif, N.}, \bibinfo{author}{Luqman, H.}, \bibinfo{author}{Ahmed, M.}, \bibinfo{year}{2023}a.
\newblock \bibinfo{title}{Enhancing stance detection through sequential weighted multi-task learning}.
\newblock \bibinfo{journal}{Social Network Analysis and Mining} \bibinfo{volume}{14}, \bibinfo{pages}{7}.
\bibitem[{Alturayeif et~al.(2023b)Alturayeif, Luqman and Ahmed}]{alturayeif2023systematic}
\bibinfo{author}{Alturayeif, N.}, \bibinfo{author}{Luqman, H.}, \bibinfo{author}{Ahmed, M.}, \bibinfo{year}{2023}b.
\newblock \bibinfo{title}{A systematic review of machine learning techniques for stance detection and its applications}.
\newblock \bibinfo{journal}{Neural Computing and Applications} \bibinfo{volume}{35}, \bibinfo{pages}{5113--5144}.
\bibitem[{Alturayeif et~al.(2024)Alturayeif, Luqman and Ahmed}]{alturayeif2024correction}
\bibinfo{author}{Alturayeif, N.}, \bibinfo{author}{Luqman, H.}, \bibinfo{author}{Ahmed, M.}, \bibinfo{year}{2024}.
\newblock \bibinfo{title}{Correction: Enhancing stance detection through sequential weighted multi-task learning}.
\newblock \bibinfo{journal}{Social Network Analysis and Mining} \bibinfo{volume}{14}, \bibinfo{pages}{25}.
\bibitem[{Augenstein et~al.(2016)Augenstein, Rockt{\"a}schel, Vlachos and Bontcheva}]{augenstein2016stance}
\bibinfo{author}{Augenstein, I.}, \bibinfo{author}{Rockt{\"a}schel, T.}, \bibinfo{author}{Vlachos, A.}, \bibinfo{author}{Bontcheva, K.}, \bibinfo{year}{2016}.
\newblock \bibinfo{title}{Stance detection with bidirectional conditional encoding}.
\newblock \bibinfo{journal}{arXiv preprint arXiv:1606.05464} .
\bibitem[{Conforti et~al.(2020)Conforti, Berndt, Pilehvar, Giannitsarou, Toxvaerd and Collier}]{conforti2020stander}
\bibinfo{author}{Conforti, C.}, \bibinfo{author}{Berndt, J.}, \bibinfo{author}{Pilehvar, M.T.}, \bibinfo{author}{Giannitsarou, C.}, \bibinfo{author}{Toxvaerd, F.}, \bibinfo{author}{Collier, N.}, \bibinfo{year}{2020}.
\newblock \bibinfo{title}{Stander: an expert-annotated dataset for news stance detection and evidence retrieval}, \bibinfo{organization}{Association for Computational Linguistics}.
\bibitem[{Du et~al.(2017)Du, Xu, He and Gui}]{du2017stance}
\bibinfo{author}{Du, J.}, \bibinfo{author}{Xu, R.}, \bibinfo{author}{He, Y.}, \bibinfo{author}{Gui, L.}, \bibinfo{year}{2017}.
\newblock \bibinfo{title}{Stance classification with target-specific neural attention networks}, \bibinfo{organization}{International Joint Conferences on Artificial Intelligence}.
\bibitem[{Ferreyra et~al.(2022)Ferreyra, Hecking, A{\"\i}meur, Heisel and Hoppe}]{ferreyra2022community}
\bibinfo{author}{Ferreyra, N.E.D.}, \bibinfo{author}{Hecking, T.}, \bibinfo{author}{A{\"\i}meur, E.}, \bibinfo{author}{Heisel, M.}, \bibinfo{author}{Hoppe, H.U.}, \bibinfo{year}{2022}.
\newblock \bibinfo{title}{Community detection for access-control decisions: Analysing the role of homophily and information diffusion in online social networks}.
\newblock \bibinfo{journal}{Online Social Networks and Media} \bibinfo{volume}{29}, \bibinfo{pages}{100203}.
\bibitem[{Goyal and Ferrara(2018)}]{goyal2018graph}
\bibinfo{author}{Goyal, P.}, \bibinfo{author}{Ferrara, E.}, \bibinfo{year}{2018}.
\newblock \bibinfo{title}{Graph embedding techniques, applications, and performance: A survey}.
\newblock \bibinfo{journal}{Knowledge-Based Systems} \bibinfo{volume}{151}, \bibinfo{pages}{78--94}.
\bibitem[{Hanselowski et~al.(2018)Hanselowski, PVS, Schiller, Caspelherr, Chaudhuri, Meyer and Gurevych}]{hanselowski2018retrospective}
\bibinfo{author}{Hanselowski, A.}, \bibinfo{author}{PVS, A.}, \bibinfo{author}{Schiller, B.}, \bibinfo{author}{Caspelherr, F.}, \bibinfo{author}{Chaudhuri, D.}, \bibinfo{author}{Meyer, C.M.}, \bibinfo{author}{Gurevych, I.}, \bibinfo{year}{2018}.
\newblock \bibinfo{title}{A retrospective analysis of the fake news challenge stance detection task}.
\newblock \bibinfo{journal}{arXiv preprint arXiv:1806.05180} .
\bibitem[{Hasan and Ng(2013)}]{hasan2013stance}
\bibinfo{author}{Hasan, K.S.}, \bibinfo{author}{Ng, V.}, \bibinfo{year}{2013}.
\newblock \bibinfo{title}{Stance classification of ideological debates: Data, models, features, and constraints}, in: \bibinfo{booktitle}{Proceedings of the sixth international joint conference on natural language processing}, pp. \bibinfo{pages}{1348--1356}.
\bibitem[{He et~al.(2022)He, Mokhberian and Lerman}]{he2022infusing}
\bibinfo{author}{He, Z.}, \bibinfo{author}{Mokhberian, N.}, \bibinfo{author}{Lerman, K.}, \bibinfo{year}{2022}.
\newblock \bibinfo{title}{Infusing knowledge from wikipedia to enhance stance detection}.
\newblock \bibinfo{journal}{arXiv preprint arXiv:2204.03839} .
\bibitem[{Jiang et~al.(2022)Jiang, Gao, Shen and Cheng}]{jiang2022few}
\bibinfo{author}{Jiang, Y.}, \bibinfo{author}{Gao, J.}, \bibinfo{author}{Shen, H.}, \bibinfo{author}{Cheng, X.}, \bibinfo{year}{2022}.
\newblock \bibinfo{title}{Few-shot stance detection via target-aware prompt distillation}, in: \bibinfo{booktitle}{Proceedings of the 45th International ACM SIGIR Conference on Research and Development in Information Retrieval}, pp. \bibinfo{pages}{837--847}.
\bibitem[{Khiabani and Zubiaga(2023)}]{khiabani2023few}
\bibinfo{author}{Khiabani, P.J.}, \bibinfo{author}{Zubiaga, A.}, \bibinfo{year}{2023}.
\newblock \bibinfo{title}{Few-shot learning for cross-target stance detection by aggregating multimodal embeddings}.
\newblock \bibinfo{journal}{IEEE Transactions on Computational Social Systems} .
\bibitem[{Kochkina et~al.(2023)Kochkina, Hossain, Logan~IV, Arana-Catania, Procter, Zubiaga, Singh, He and Liakata}]{kochkina2023evaluating}
\bibinfo{author}{Kochkina, E.}, \bibinfo{author}{Hossain, T.}, \bibinfo{author}{Logan~IV, R.L.}, \bibinfo{author}{Arana-Catania, M.}, \bibinfo{author}{Procter, R.}, \bibinfo{author}{Zubiaga, A.}, \bibinfo{author}{Singh, S.}, \bibinfo{author}{He, Y.}, \bibinfo{author}{Liakata, M.}, \bibinfo{year}{2023}.
\newblock \bibinfo{title}{Evaluating the generalisability of neural rumour verification models}.
\newblock \bibinfo{journal}{Information Processing \& Management} \bibinfo{volume}{60}, \bibinfo{pages}{103116}.
\bibitem[{K{\"u}{\c{c}}{\"u}k and Can(2020)}]{kuccuk2020stance}
\bibinfo{author}{K{\"u}{\c{c}}{\"u}k, D.}, \bibinfo{author}{Can, F.}, \bibinfo{year}{2020}.
\newblock \bibinfo{title}{Stance detection: A survey}.
\newblock \bibinfo{journal}{ACM Computing Surveys (CSUR)} \bibinfo{volume}{53}, \bibinfo{pages}{1--37}.
\bibitem[{Li et~al.(2024)Li, Zhao, Liang, Gui, Wang, Zeng, Wong and Xu}]{li2024mitigating}
\bibinfo{author}{Li, A.}, \bibinfo{author}{Zhao, J.}, \bibinfo{author}{Liang, B.}, \bibinfo{author}{Gui, L.}, \bibinfo{author}{Wang, H.}, \bibinfo{author}{Zeng, X.}, \bibinfo{author}{Wong, K.F.}, \bibinfo{author}{Xu, R.}, \bibinfo{year}{2024}.
\newblock \bibinfo{title}{Mitigating biases of large language models in stance detection with calibration}.
\newblock \bibinfo{journal}{arXiv preprint arXiv:2402.14296} .
\bibitem[{Li et~al.(2023a)Li, Garg and Caragea}]{li2023new}
\bibinfo{author}{Li, Y.}, \bibinfo{author}{Garg, K.}, \bibinfo{author}{Caragea, C.}, \bibinfo{year}{2023}a.
\newblock \bibinfo{title}{A new direction in stance detection: Target-stance extraction in the wild}, in: \bibinfo{booktitle}{Proceedings of the 61st Annual Meeting of the Association for Computational Linguistics (Volume 1: Long Papers)}, pp. \bibinfo{pages}{10071--10085}.
\bibitem[{Li et~al.(2023b)Li, He, Wang, Lau and Song}]{li2023improved}
\bibinfo{author}{Li, Y.}, \bibinfo{author}{He, H.}, \bibinfo{author}{Wang, S.}, \bibinfo{author}{Lau, F.C.}, \bibinfo{author}{Song, Y.}, \bibinfo{year}{2023}b.
\newblock \bibinfo{title}{Improved target-specific stance detection on social media platforms by delving into conversation threads}.
\newblock \bibinfo{journal}{IEEE Transactions on Computational Social Systems} .
\bibitem[{Li et~al.(2021)Li, Sosea, Sawant, Nair, Inkpen and Caragea}]{li2021p}
\bibinfo{author}{Li, Y.}, \bibinfo{author}{Sosea, T.}, \bibinfo{author}{Sawant, A.}, \bibinfo{author}{Nair, A.J.}, \bibinfo{author}{Inkpen, D.}, \bibinfo{author}{Caragea, C.}, \bibinfo{year}{2021}.
\newblock \bibinfo{title}{P-stance: A large dataset for stance detection in political domain}, in: \bibinfo{booktitle}{Findings of the Association for Computational Linguistics: ACL-IJCNLP 2021}, pp. \bibinfo{pages}{2355--2365}.
\bibitem[{Liang et~al.(2022)Liang, Zhu, Li, Yang, Gui, He and Xu}]{liang2022jointcl}
\bibinfo{author}{Liang, B.}, \bibinfo{author}{Zhu, Q.}, \bibinfo{author}{Li, X.}, \bibinfo{author}{Yang, M.}, \bibinfo{author}{Gui, L.}, \bibinfo{author}{He, Y.}, \bibinfo{author}{Xu, R.}, \bibinfo{year}{2022}.
\newblock \bibinfo{title}{Jointcl: a joint contrastive learning framework for zero-shot stance detection}, in: \bibinfo{booktitle}{Proceedings of the 60th Annual Meeting of the Association for Computational Linguistics (Volume 1: Long Papers)}, \bibinfo{organization}{Association for Computational Linguistics}. pp. \bibinfo{pages}{81--91}.
\bibitem[{Liu et~al.(2019)Liu, Ott, Goyal, Du, Joshi, Chen, Levy, Lewis, Zettlemoyer and Stoyanov}]{liu2019roberta}
\bibinfo{author}{Liu, Y.}, \bibinfo{author}{Ott, M.}, \bibinfo{author}{Goyal, N.}, \bibinfo{author}{Du, J.}, \bibinfo{author}{Joshi, M.}, \bibinfo{author}{Chen, D.}, \bibinfo{author}{Levy, O.}, \bibinfo{author}{Lewis, M.}, \bibinfo{author}{Zettlemoyer, L.}, \bibinfo{author}{Stoyanov, V.}, \bibinfo{year}{2019}.
\newblock \bibinfo{title}{Roberta: A robustly optimized bert pretraining approach}.
\newblock \bibinfo{journal}{arXiv preprint arXiv:1907.11692} .
\bibitem[{McPherson et~al.(2001)McPherson, Smith-Lovin and Cook}]{mcpherson2001birds}
\bibinfo{author}{McPherson, M.}, \bibinfo{author}{Smith-Lovin, L.}, \bibinfo{author}{Cook, J.M.}, \bibinfo{year}{2001}.
\newblock \bibinfo{title}{Birds of a feather: Homophily in social networks}.
\newblock \bibinfo{journal}{Annual review of sociology} \bibinfo{volume}{27}, \bibinfo{pages}{415--444}.
\bibitem[{Medhat et~al.(2014)Medhat, Hassan and Korashy}]{medhat2014sentiment}
\bibinfo{author}{Medhat, W.}, \bibinfo{author}{Hassan, A.}, \bibinfo{author}{Korashy, H.}, \bibinfo{year}{2014}.
\newblock \bibinfo{title}{Sentiment analysis algorithms and applications: A survey}.
\newblock \bibinfo{journal}{Ain Shams engineering journal} \bibinfo{volume}{5}, \bibinfo{pages}{1093--1113}.
\bibitem[{Mohammad et~al.(2016)Mohammad, Kiritchenko, Sobhani, Zhu and Cherry}]{mohammad2016semeval}
\bibinfo{author}{Mohammad, S.}, \bibinfo{author}{Kiritchenko, S.}, \bibinfo{author}{Sobhani, P.}, \bibinfo{author}{Zhu, X.}, \bibinfo{author}{Cherry, C.}, \bibinfo{year}{2016}.
\newblock \bibinfo{title}{Semeval-2016 task 6: Detecting stance in tweets}, in: \bibinfo{booktitle}{Proceedings of the 10th international workshop on semantic evaluation (SemEval-2016)}, pp. \bibinfo{pages}{31--41}.
\bibitem[{Mohtarami et~al.(2018)Mohtarami, Baly, Glass, Nakov, M{\`a}rquez and Moschitti}]{mohtarami2018automatic}
\bibinfo{author}{Mohtarami, M.}, \bibinfo{author}{Baly, R.}, \bibinfo{author}{Glass, J.}, \bibinfo{author}{Nakov, P.}, \bibinfo{author}{M{\`a}rquez, L.}, \bibinfo{author}{Moschitti, A.}, \bibinfo{year}{2018}.
\newblock \bibinfo{title}{Automatic stance detection using end-to-end memory networks}.
\newblock \bibinfo{journal}{arXiv preprint arXiv:1804.07581} .
\bibitem[{Ostendorff et~al.(2019)Ostendorff, Bourgonje, Berger, Moreno-Schneider, Rehm and Gipp}]{ostendorff2019enriching}
\bibinfo{author}{Ostendorff, M.}, \bibinfo{author}{Bourgonje, P.}, \bibinfo{author}{Berger, M.}, \bibinfo{author}{Moreno-Schneider, J.}, \bibinfo{author}{Rehm, G.}, \bibinfo{author}{Gipp, B.}, \bibinfo{year}{2019}.
\newblock \bibinfo{title}{Enriching bert with knowledge graph embeddings for document classification}.
\newblock \bibinfo{journal}{arXiv preprint arXiv:1909.08402} .
\bibitem[{Riedel et~al.(2017)Riedel, Augenstein, Spithourakis and Riedel}]{riedel2017simple}
\bibinfo{author}{Riedel, B.}, \bibinfo{author}{Augenstein, I.}, \bibinfo{author}{Spithourakis, G.P.}, \bibinfo{author}{Riedel, S.}, \bibinfo{year}{2017}.
\newblock \bibinfo{title}{A simple but tough-to-beat baseline for the fake news challenge stance detection task}.
\newblock \bibinfo{journal}{arXiv preprint arXiv:1707.03264} .
\bibitem[{S{\'a}enz and Becker(2021)}]{saenz2021interpreting}
\bibinfo{author}{S{\'a}enz, C.A.C.}, \bibinfo{author}{Becker, K.}, \bibinfo{year}{2021}.
\newblock \bibinfo{title}{Interpreting bert-based stance classification: a case study about the brazilian covid vaccination}, in: \bibinfo{booktitle}{Anais do XXXVI Simp{\'o}sio Brasileiro de Bancos de Dados}, \bibinfo{organization}{SBC}. pp. \bibinfo{pages}{73--84}.
\bibitem[{Schick and Sch{\"u}tze(2020)}]{schick2020exploiting}
\bibinfo{author}{Schick, T.}, \bibinfo{author}{Sch{\"u}tze, H.}, \bibinfo{year}{2020}.
\newblock \bibinfo{title}{Exploiting cloze questions for few shot text classification and natural language inference}.
\newblock \bibinfo{journal}{arXiv preprint arXiv:2001.07676} .
\bibitem[{Shu et~al.(2017)Shu, Sliva, Wang, Tang and Liu}]{shu2017fake}
\bibinfo{author}{Shu, K.}, \bibinfo{author}{Sliva, A.}, \bibinfo{author}{Wang, S.}, \bibinfo{author}{Tang, J.}, \bibinfo{author}{Liu, H.}, \bibinfo{year}{2017}.
\newblock \bibinfo{title}{Fake news detection on social media: A data mining perspective}.
\newblock \bibinfo{journal}{ACM SIGKDD explorations newsletter} \bibinfo{volume}{19}, \bibinfo{pages}{22--36}.
\bibitem[{Sobhani et~al.(2017)Sobhani, Inkpen and Zhu}]{sobhani2017dataset}
\bibinfo{author}{Sobhani, P.}, \bibinfo{author}{Inkpen, D.}, \bibinfo{author}{Zhu, X.}, \bibinfo{year}{2017}.
\newblock \bibinfo{title}{A dataset for multi-target stance detection}, in: \bibinfo{booktitle}{Proceedings of the 15th Conference of the European Chapter of the Association for Computational Linguistics: Volume 2, Short Papers}, pp. \bibinfo{pages}{551--557}.
\bibitem[{Sridhar et~al.(2015)Sridhar, Foulds, Huang, Getoor and Walker}]{sridhar2015joint}
\bibinfo{author}{Sridhar, D.}, \bibinfo{author}{Foulds, J.}, \bibinfo{author}{Huang, B.}, \bibinfo{author}{Getoor, L.}, \bibinfo{author}{Walker, M.}, \bibinfo{year}{2015}.
\newblock \bibinfo{title}{Joint models of disagreement and stance in online debate}, in: \bibinfo{booktitle}{Proceedings of the 53rd Annual Meeting of the Association for Computational Linguistics and the 7th International Joint Conference on Natural Language Processing (Volume 1: Long Papers)}, pp. \bibinfo{pages}{116--125}.
\bibitem[{Sun et~al.(2018)Sun, Wang, Zhu and Zhou}]{sun2018stance}
\bibinfo{author}{Sun, Q.}, \bibinfo{author}{Wang, Z.}, \bibinfo{author}{Zhu, Q.}, \bibinfo{author}{Zhou, G.}, \bibinfo{year}{2018}.
\newblock \bibinfo{title}{Stance detection with hierarchical attention network}, in: \bibinfo{booktitle}{Proceedings of the 27th international conference on computational linguistics}, pp. \bibinfo{pages}{2399--2409}.
\bibitem[{Sun et~al.(2017)Sun, Luo and Chen}]{sun2017review}
\bibinfo{author}{Sun, S.}, \bibinfo{author}{Luo, C.}, \bibinfo{author}{Chen, J.}, \bibinfo{year}{2017}.
\newblock \bibinfo{title}{A review of natural language processing techniques for opinion mining systems}.
\newblock \bibinfo{journal}{Information fusion} \bibinfo{volume}{36}, \bibinfo{pages}{10--25}.
\bibitem[{Wen and Hauptmann(2023)}]{wen2023zero}
\bibinfo{author}{Wen, H.}, \bibinfo{author}{Hauptmann, A.G.}, \bibinfo{year}{2023}.
\newblock \bibinfo{title}{Zero-shot and few-shot stance detection on varied topics via conditional generation}, in: \bibinfo{booktitle}{Proceedings of the 61st Annual Meeting of the Association for Computational Linguistics (Volume 2: Short Papers)}, pp. \bibinfo{pages}{1491--1499}.
\bibitem[{Xu et~al.(2018)Xu, Paris, Nepal and Sparks}]{xu2018cross}
\bibinfo{author}{Xu, C.}, \bibinfo{author}{Paris, C.}, \bibinfo{author}{Nepal, S.}, \bibinfo{author}{Sparks, R.}, \bibinfo{year}{2018}.
\newblock \bibinfo{title}{Cross-target stance classification with self-attention networks}.
\newblock \bibinfo{journal}{arXiv preprint arXiv:1805.06593} .
\bibitem[{Zhao et~al.(2023)Zhao, Li and Caragea}]{zhao2023c}
\bibinfo{author}{Zhao, C.}, \bibinfo{author}{Li, Y.}, \bibinfo{author}{Caragea, C.}, \bibinfo{year}{2023}.
\newblock \bibinfo{title}{C-stance: A large dataset for chinese zero-shot stance detection}, in: \bibinfo{booktitle}{Proceedings of the 61st Annual Meeting of the Association for Computational Linguistics (Volume 1: Long Papers)}, pp. \bibinfo{pages}{13369--13385}.
\bibitem[{Zubiaga(2018)}]{zubiaga2018longitudinal}
\bibinfo{author}{Zubiaga, A.}, \bibinfo{year}{2018}.
\newblock \bibinfo{title}{A longitudinal assessment of the persistence of twitter datasets}.
\newblock \bibinfo{journal}{Journal of the Association for Information Science and Technology} \bibinfo{volume}{69}, \bibinfo{pages}{974--984}.
\bibitem[{Zubiaga(2024)}]{zubiaga2024natural}
\bibinfo{author}{Zubiaga, A.}, \bibinfo{year}{2024}.
\newblock \bibinfo{title}{Natural language processing in the era of large language models}.
\newblock \bibinfo{journal}{Frontiers in Artificial Intelligence} \bibinfo{volume}{6}, \bibinfo{pages}{1350306}.
\bibitem[{Zubiaga et~al.(2016)Zubiaga, Kochkina, Liakata, Procter and Lukasik}]{zubiaga2016stance}
\bibinfo{author}{Zubiaga, A.}, \bibinfo{author}{Kochkina, E.}, \bibinfo{author}{Liakata, M.}, \bibinfo{author}{Procter, R.}, \bibinfo{author}{Lukasik, M.}, \bibinfo{year}{2016}.
\newblock \bibinfo{title}{Stance classification in rumours as a sequential task exploiting the tree structure of social media conversations}.
\newblock \bibinfo{journal}{arXiv preprint arXiv:1609.09028} .
\bibitem[{Zubiaga et~al.(2019)Zubiaga, Wang, Liakata and Procter}]{zubiaga2019political}
\bibinfo{author}{Zubiaga, A.}, \bibinfo{author}{Wang, B.}, \bibinfo{author}{Liakata, M.}, \bibinfo{author}{Procter, R.}, \bibinfo{year}{2019}.
\newblock \bibinfo{title}{Political homophily in independence movements: analyzing and classifying social media users by national identity}.
\newblock \bibinfo{journal}{IEEE Intelligent Systems} \bibinfo{volume}{34}, \bibinfo{pages}{34--42}.

\end{thebibliography}





\end{document}